\documentclass[10pt,twocolumn,letterpaper]{article}

\usepackage{booktabs}
\usepackage{authblk}
\usepackage{cvpr}
\usepackage{times}
\usepackage{epsfig}
\usepackage{graphicx}
\usepackage{amsmath}
\usepackage{amssymb}
\usepackage[mathscr]{eucal}
\usepackage{bm,url} 
\usepackage[tight,large]{subfigure}

\usepackage[pagebackref=true,breaklinks=true,letterpaper=true,colorlinks,bookmarks=false]{hyperref}

\cvprfinalcopy 

\def\cvprPaperID{29} 

\renewcommand{\b}{\mathbf}

\usepackage[usenames,dvipsnames]{color}
\usepackage[normalem]{ulem}

\ifcvprfinal\fi
\begin{document}

\title{Emotional Expression Classification using Time-Series Kernels\thanks{\copyright{}2013 IEEE. IEEE International Workshop on Analysis and Modeling of Faces and Gestures, Portland, Oregon, 28 June 2013 (accepted).}}

\author[1]{Andr\'as L\H orincz}
\author[2]{L\'aszl\'o A. Jeni}
\author[1]{Zolt\'an Szab\'o}
\author[2, 3]{Jeffrey F. Cohn}
\author[2]{Takeo Kanade}

\affil[1]{E\"otv\"os Lor\'and University\\
Budapest, Hungary\\
\url{{andras.lorincz, szzoli}@elte.hu}}
\affil[2]{Carnegie Mellon University\\
Pittsburgh, PA\\
\url{laszlo.jeni@ieee.org, tk@cs.cmu.edu}}
\affil[3]{University of Pittsburgh\\
Pittsburgh, PA\\
\url{jeffcohn@cs.cmu.edu}}

\maketitle

\begin{abstract}
Estimation of facial expressions, as spatio-temporal processes, can take advantage of kernel methods if one considers facial landmark positions and their motion in 3D space. We applied support vector classification with kernels derived from dynamic time-warping similarity measures. We achieved over 99\% accuracy --~measured by area under ROC curve~-- using only the 'motion pattern' of the PCA compressed representation of the marker point vector, the so-called shape parameters. Beyond the classification of full motion patterns, several expressions were recognized with over 90\% accuracy in as few as 5-6 frames from their onset, about 200 milliseconds.
\end{abstract}

\section{Introduction}

Because they enable model based prediction and timely reactions, the analysis and identification of spatio-temporal processes, including concurrent and interacting events, are of great importance in many applications. Spatio-temporal processes have intriguing features. Consider three examples. First is a feature-length movie, which is a set of time series of pixel intensities in which the number of pixels is on the order of 100,000 and the values follow each other at a rate of 30 fps for a time interval of 2 hours or so. Second are financial time series. Currency exchange rates, stocks, and many other finance related data are heavily affected by both common underlying processes and by one another and exhibit large coupled fluctuations over 9 orders of magnitude \cite{preis2011switching}. Third, and a focus of the current paper, are facial expression time series. The motion patterns of landmark points of a face, such as mouth and eye corners comprise a 'landmark space'. The dynamics of change in this space can reveal emotion, pain, and cognitive states, and regulate social interaction. For effective human-computer interaction, automated facial expression analysis is important.

\begin{figure}[!h]
\centering
\includegraphics[width=7cm]{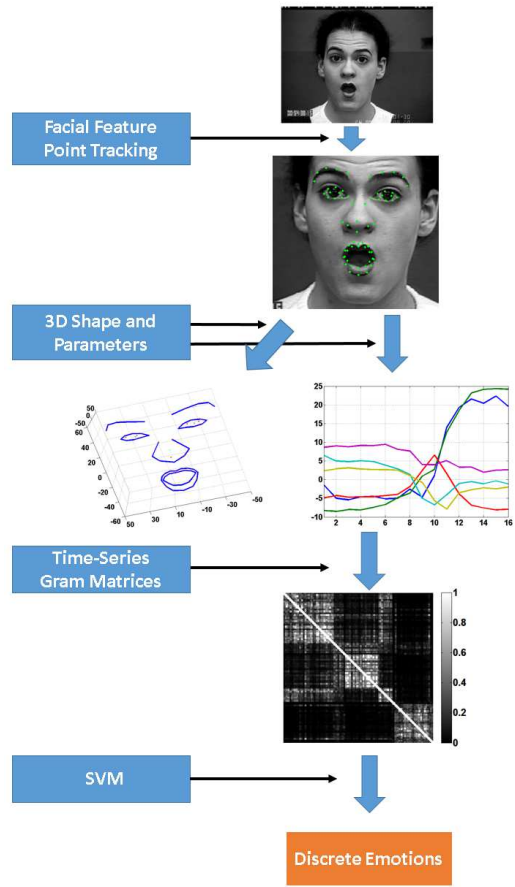}
\caption[]{Overview of the system. }
\label{fig:overview}
\end{figure}

In all three types of time series, i.e., movies, financial data, and facial expressions, we are dealing with spatio-temporal processes. For each, kernel methods hold great promise.

The demands for characterizing such processes pose special challenges because very different signals may represent the same process when the process is viewed from different distances (in the case of the movie), different time scales (market data), or different viewing angles (facial expressions). Such invariances and distortions need be taken into account. 

Early attempts to use support vector methods for the prediction of time series were very promising \cite{muller97predicting} even in the absence of algorithms compensating for temporal distortions. In other areas, such as speech recognition, time warping algorithms have been developed early in order to match slower and faster speech fragments, see, e.g., \cite{deller2000discrete} and the references therein. Dynamic time warping is one of the most efficient methods that offer the comparison of temporally distorted samples \cite{sakoe1978dynamic}. 

Recently, the two methods, i.e., dynamic time warping and SVMs have been combined and show considerable performance increases in the analysis of spatio-temporal signals \cite{cuturi07akernel,cuturi2011fast}. Efficient methods using independent component analysis \cite{Long2012126}, Haar filters \cite{yang2009boosting}, hidden Markov models \cite{BousmalisMP11,BousmalisZMP13,valstar2007combined} have been applied for problems related to the estimation of emotions and facial expressions. Here, we study the efficiency of novel dynamic time warping kernels \cite{cuturi07akernel,cuturi2011fast} for emotional expression estimation.

Our contributions are as follows: We show that (1) time series kernel methods are highly precise for emotional expression estimation using landmark data only and (2) they enable early and reliable estimation of expression as soon as 5 frames from expression onset, i.e., around 200 ms.

The paper is organized as follows. First, in the Methods section we review how landmark points are observed in 3 dimensions, sketch the two spatio-temporal kernels that we applied, and describe support vector machine (SVM) principles. Section~\ref{s:exps} is about our experimental studies. It is followed by our Discussion and Summary.
 
\section{Methods}

\subsection{Facial Feature Point Localization}

To localize a dense set of facial landmarks, Active Appearance Models (AAM) \cite{matthews2004active} and Constrained Local Models (CLM) \cite{saragih2011deformable} are often used. These methods register a dense  parameterized shape model to an image such that its landmarks correspond to consistent locations on the face. 

Of the two, person specific AAMs have higher precision than CLMs, but they must be trained for each person before use. On the other hand, CLM methods can be used for person-independent face alignment because of the localized region templates.

In this work we use a 3D CLM method, where the shape model is defined by a 3D mesh and in particular the 3D vertex locations of the mesh, called landmark points. Consider the shape of a 3D CLM as the coordinates of 3D vertices that make up the mesh:

\begin{equation}
\b{x} = [ x_1; y_1; z_1; \ldots ; x_M; y_M; z_M ],
\end{equation}

or, $\b{x}=[\b{x}_1; \ldots ; \b{x}_M]$, where $\b{x}_i=[x_i;y_i;z_i]$. We have $T$ samples: $\{\b{x}(t)\}_{t=1}^T$.
We assume that -- apart from scale, rotation, and translation -- all samples $\{\b{x}(t)\}_{t=1}^T$ can be approximated by means of the linear principal component analysis (PCA).

In the next subsection we briefly describe the 3D Point Distribution Model and how the CLM method estimates the landmark positions.

\subsubsection{Point Distribution Model}
The 3D point distribution model (PDM) describes non-rigid shape variations linearly and composes it with a global rigid transformation, placing the shape in the image frame:

\begin{equation}\label{eq:PDM}
\b{x}_i(p)=s\b{PR}(\bar{\b{x}}_i+\bm{\Phi}_i\b{q})+\b{t},
\end{equation}

$(i=1,\ldots,M)$, where $\b{x}_i(p)$ denotes the 3D location of the $i^{th}$ landmark and $\b{p} = \{s,\alpha,\beta,\gamma,\b{q},\b{t}\}$ denotes the parameters of the model, which consist of a global scaling $s$, angles of rotation in three dimensions ($\b{R}= \b{R}_1(\alpha)\b{R}_2(\beta)\b{R}_3(\gamma)$), a translation $\b{t}$ and non-rigid transformation $\b{q}$. Here $\bar{\b{x}}_i$ denotes the mean location of the $i^{th}$ landmark (i.e. $\bar{\b{x}}_i=[\bar{x}_i; \bar{y}_i; \bar{z}_i]$ and $\bar{\b{x}}=[\bar{\b{x}}_1;\ldots;\bar{\b{x}}_M])$ and $\b{P}$ denotes the projection matrix to 2D:

\begin{align}
\b{P} &=
\left[
\begin{array}{ccc}
     1 &  0 & 0  \\
     0 &   1 & 0  \\
\end{array}
\right].
\end{align}

We assume that the prior of the parameters follow a normal distribution with mean $\b{0}$ and variance $\bm{\Lambda}$ at a parameter vector $\b{q}$:

\begin{equation}\label{eq:PDM_prior}
p(\b{p})\propto N(\b{q};\b{0},\bm{\Lambda}),
\end{equation}

From $\b{x}_i$ points PCA provides $\bar{\b{x}}$ in \eqref{eq:PDM} and $\bm{\Lambda}$ in \eqref{eq:PDM_prior}.

\subsubsection{Constrained Local Model}
CLM is constrained through the PCA of PDM. It works with \emph{local experts}, whose opinion is considered independent and are multiplied to each other:
\begin{equation}\label{eq:J}
		J(\mathbf{p})=
		p(\mathbf{p})\prod_{i=1}^M p(l_i=1|\mathbf{x}_i(\mathbf{p}),\mathcal{I}) \rightarrow \min_{\mathbf{p}},
\end{equation}
where $l_i \in \{-1,1\}$ is a stochastic variable telling whether the $i^{th}$ marker is in its position or not, $p(l_i=1|\mathbf{x}_i(\mathbf{p}),\mathcal{I})$ is the probability that for image $\mathcal{I}$ and for marker position $\mathbf{x}_i$ (being a function of parameter $\mathbf{p}$, i.e., for $\mathbf{x}_i(\mathbf{p})$) the $i^{th}$ marker is in its position.

The interested reader is referred to \cite{saragih2011deformable} for the details of the CLM algorithm.\footnote{We used the CLM software of Saragih, which is available here \url{https://github.com/kylemcdonald/FaceTracker}.}

\begin{figure*}[!ht]
\centering
\subfigure[]{\includegraphics[height=4.5cm]{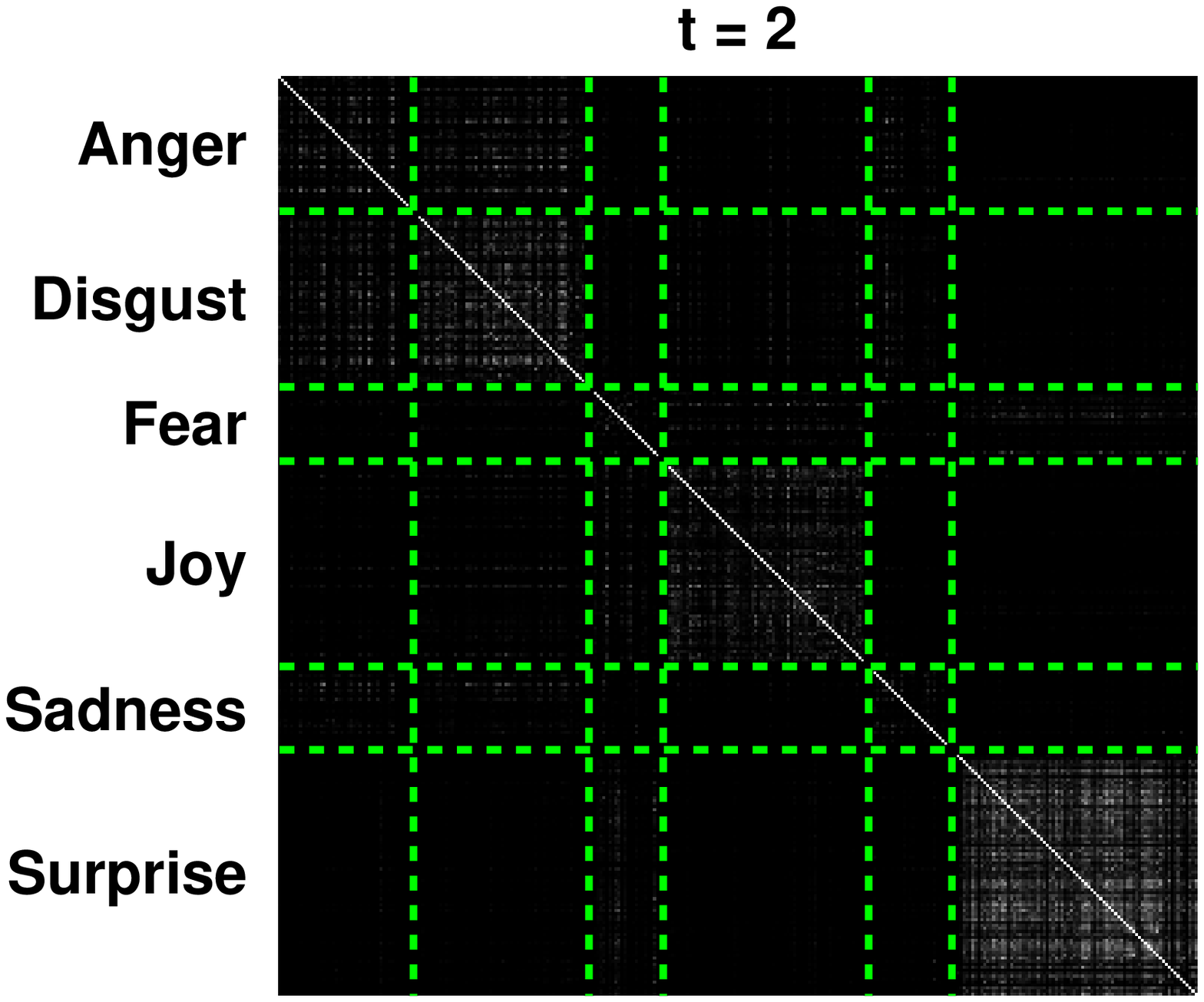}}
\subfigure[]{\includegraphics[height=4.5cm]{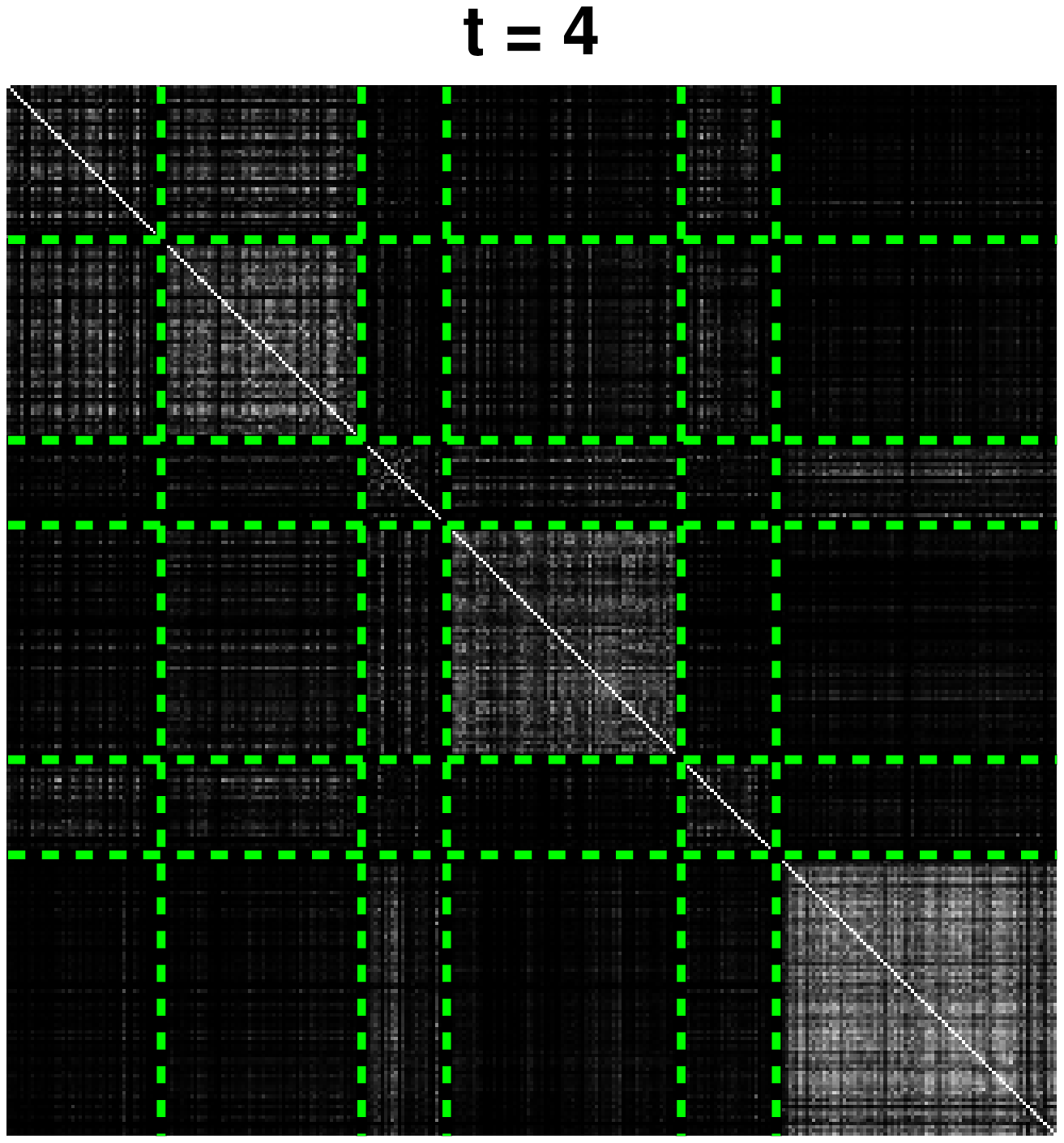}}
\subfigure[]{\includegraphics[height=4.5cm]{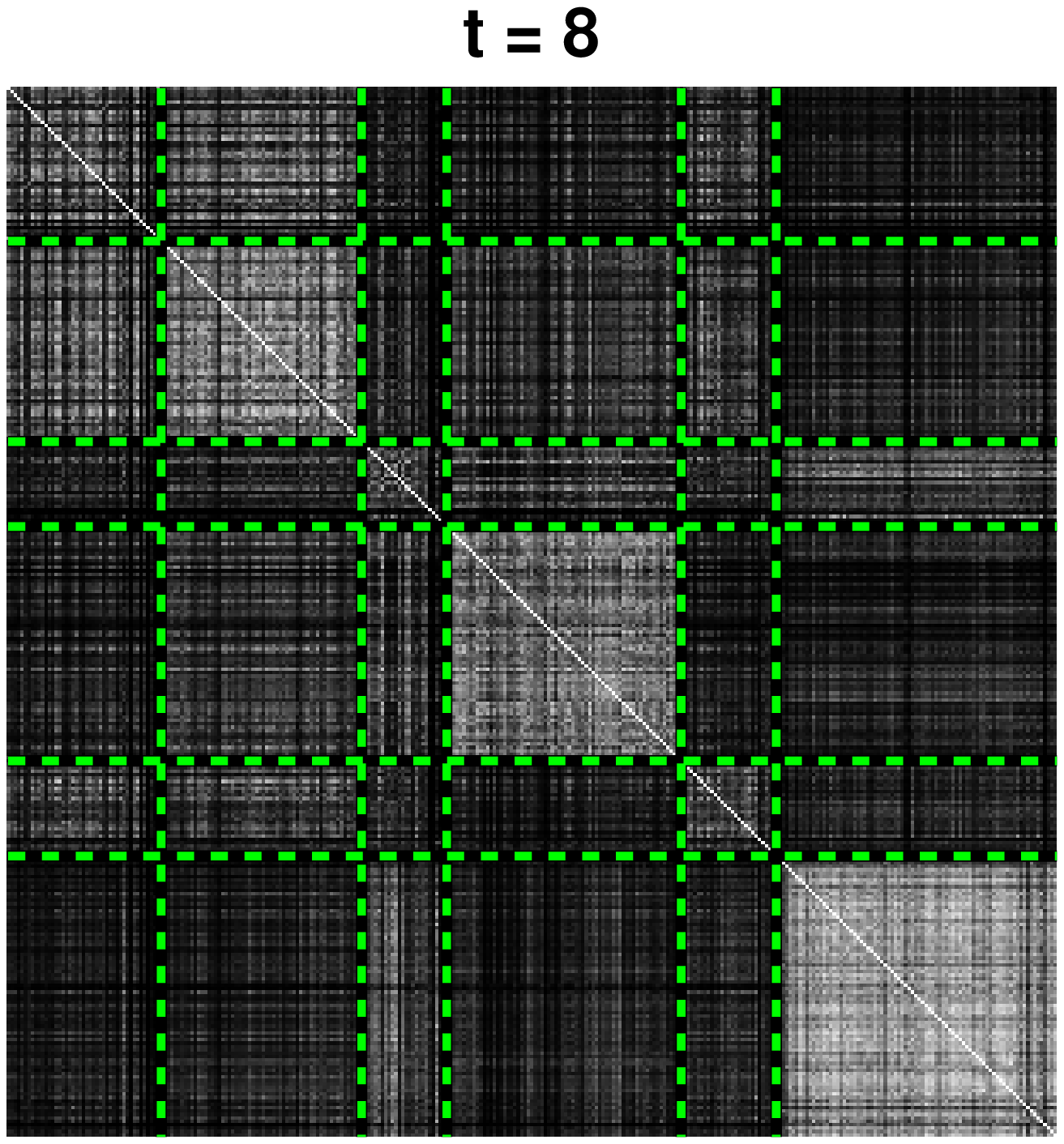}\includegraphics[height=4.5cm]{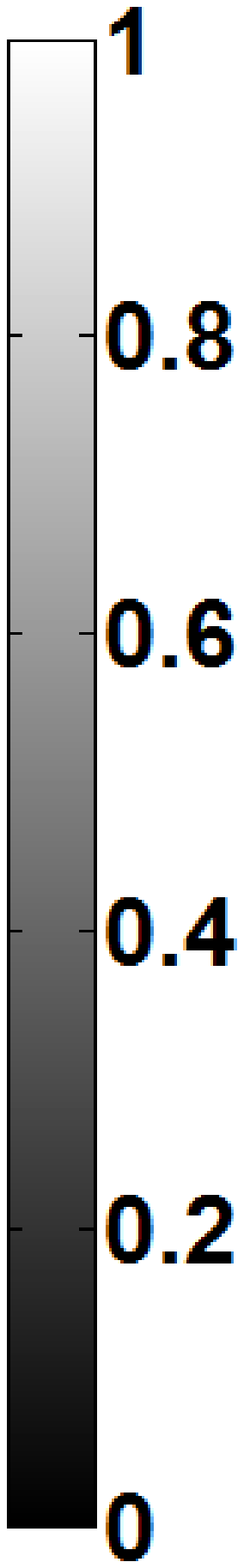}}\hfill
\subfigure[]{\includegraphics[height=4.5cm]{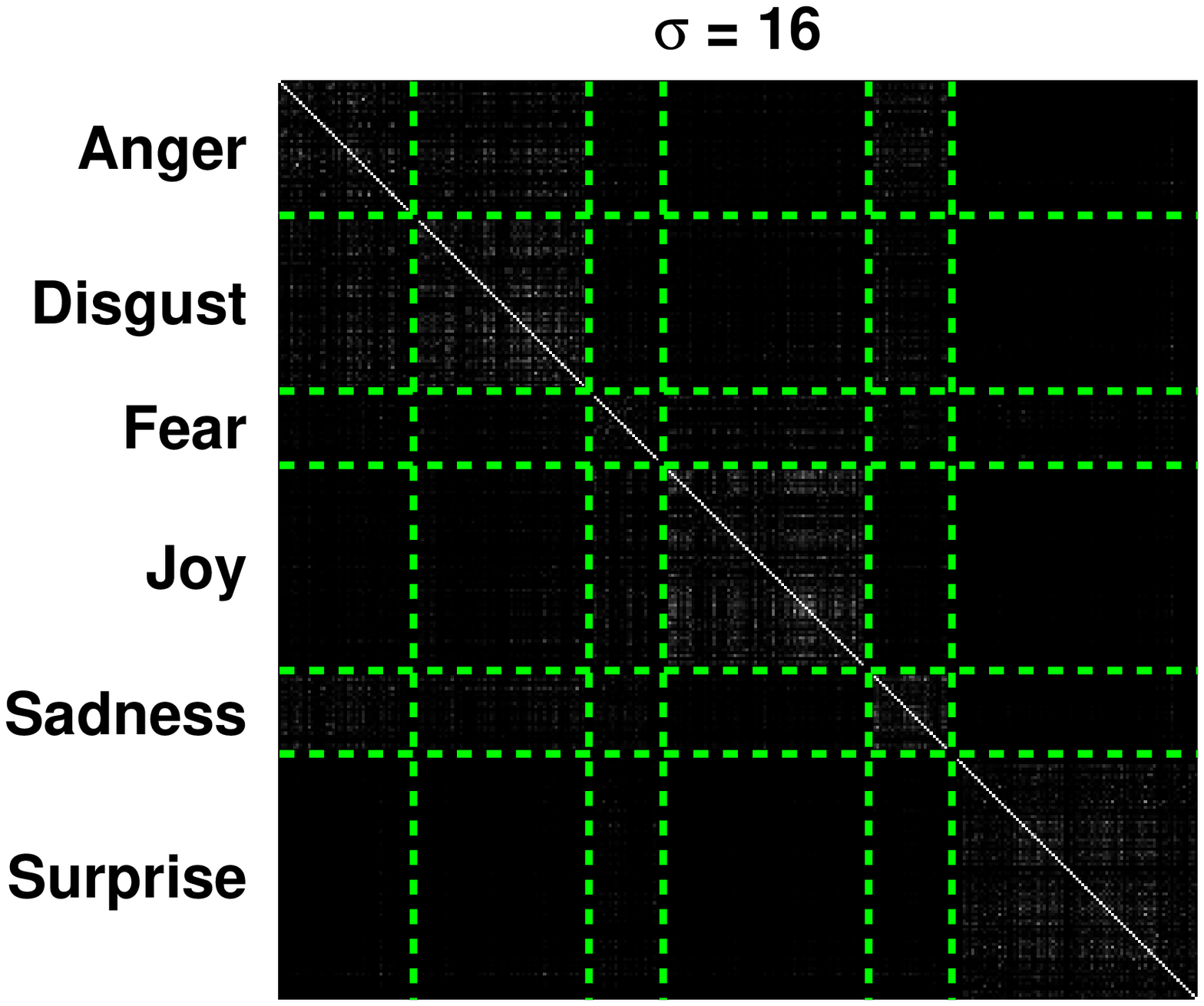}}
\subfigure[]{\includegraphics[height=4.5cm]{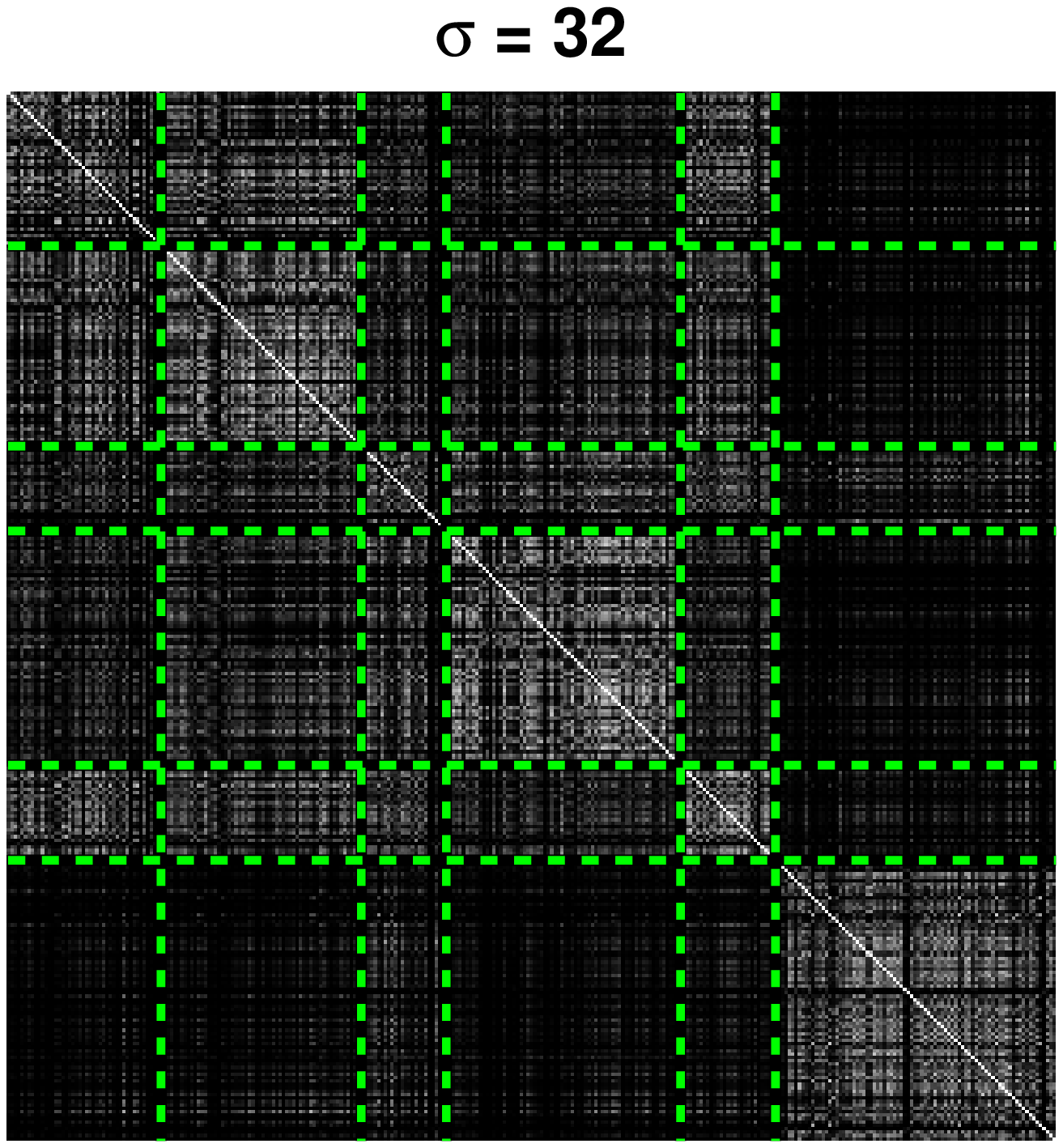}}
\subfigure[]{\includegraphics[height=4.5cm]{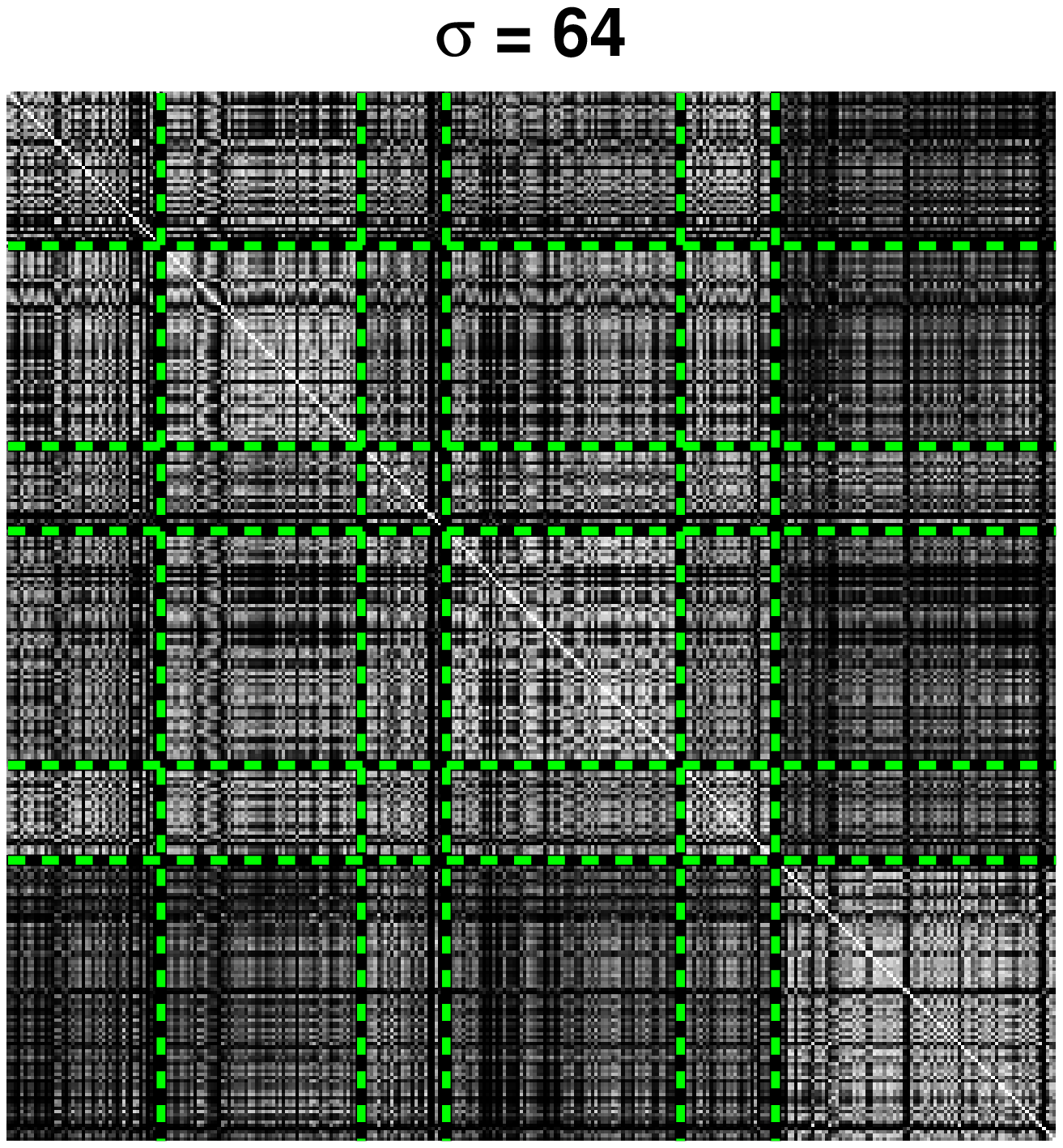}\includegraphics[height=4.5cm]{colorbar.eps}}\hfill
\caption[]{Gram matrices induced by the pseudo-DTW kernel (a-c) and the GA kernel (d-f) with different parameters. The rows and columns represent the time-series grouped by the emotion labels (the boundaries of the different emotional sets are denoted with green dashed lines). The pixel intensities in each cell show the similarity between two time-series.}
\label{fig:Gram}
\end{figure*}

\subsection{Time-series Kernels}

Kernel based classifiers, like any other classification scheme, should be robust against invariances and distortions.  Dynamic time warping, traditionally solved by dynamic programming, has been introduced to overcome temporal distortions and has been successfully combined with kernel methods. Below, we describe two kernels that we applied in our numerical studies: the Dynamic Time Warping (DTW) kernel and the Global Alignment (GA) kernel.

\subsubsection{Dynamic Time Warping Kernel}

Let $\mathscr{X}^\mathbb{N}$ be the set of discrete-time time series taking values in an arbitrary space $\mathscr{X}$. One can try to align two time series $x = (x_1, ... , x_n)$ and $y = (y_1, ... , y_m)$ of lengths $n$ and $m$, respectively, in various ways by distorting them. An alignment $\pi$ has length $p$ and $p \le n + m - 1$ since the two series have $n+m$ points and they are matched at least at one point of time. We use the notation of \cite{cuturi2011fast}. An alignment $\pi$ is a pair of increasing integral vectors $(\pi_1, \pi_2)$ of length $p$ such that $1 = \pi_1(1) \le ... \le \pi_1(p) = n$ and $1 = \pi_2(1) \le ... \le \pi_2(p) = m$, with unitary increments and no simultaneous repetitions. In turn, for all indices $1 \le i \le p-1$, the increment vector of $\pi$ belongs to a set of 3 elementary moves as follows

\begin{equation}
\left( \frac{\pi_1(i + 1)-\pi_1(i)}{\pi_2(i + 1)-\pi_2(i)} \right) \in \left\{ \left( \frac{0}{1} \right), \left( \frac{1}{0} \right), \left( \frac{1}{1} \right) \right\}
\end{equation}
Coordinates of $\pi$ are also known as warping functions. Let $A(n, m)$ denote the set of all alignments between two time series of length $n$ and $m$. The simplest DTW 'distance' between $x$ and $y$ is defined as
\begin{equation}\label{eq:DTW_def}
DTW(x,y) \overset{def}{=} \min_{\pi \in A(n,m)} D_{x,y}(\pi)
\end{equation}
Now, let $|\pi|$ denote the length of alignment $\pi$. The \emph{cost} can be defined by means of a local divergence $\phi$ that measures the discrepancy between any two points $x_i$ and $y_j$ of vectors $x$ and $y$.
\begin{equation}
D_{x,y}(\pi) \overset{def}{=} \sum_i^{|\pi|}{ \phi(x_{\pi_{1}(i)},y_{\pi_{2}(i)})}
\end{equation}
The squared Euclidean distance is often used to define the divergence $\phi(x, y) = ||x-y||^2$. Although this measure is symmetric, it does not satisfy the triangle inequality under all conditions --~so it is not rigorously a distance~-- and cannot be used directly to define a positive semi-definite kernel. This problem can be alleviated by projecting matrix $D_{x,y}(\pi)$ to a set of symmetric positive semi-definite matrices. There are various methods for accomplishing such approximations. They called distance substitution \cite{haasdonk2004learning}. We applied the alternating projection method of \cite{higham2002computing} that finds the nearest correlation matrix. Denoting the new matrix by $\hat{DTW}(x,y)$, the modified DTW distance induces a positive semi-definite kernel as follows
\begin{equation}
k_{DTW}(x,y) = e^{-\frac{1}{t}\hat{DTW}(x,y)},
\end{equation}
where $t$ is a constant. 

The full procedure can be summarized as follows: (1) take the samples, (2) compute the Euclidean distances for each sample pair, (3) build the matrix from these sample pairs, (4) find the nearest correlation matrix, (5) use it to construct a kernel, and (6) compute the Gram matrix of the support vector classification problem. Fig.~\ref{fig:Gram} (a)-(c) show Gram matrices induced by the pseudo-DTW kernel with different $t$ parameters.

\subsubsection{Global Alignment Kernel}

The Global Alignment (GA) kernel assumes that the minimum value of alignments may be sensitive to peculiarities of the time series and intends to take advantage of all alignments weighted exponentially. It is defined as the sum of exponentiated and sign changed costs of the individual alignments:

\begin{equation}\label{eq:GAK_def}
k_{GA}(x,y) \overset{def}{=} \sum_{\pi \in A(n,m)} e^{-D_{x,y}(\pi)}.
\end{equation}

Equation ~\eqref{eq:GAK_def} can be rewritten by breaking up the alignment distances according to the local divergences: similarity function $\kappa$ is induced by divergence $\phi$:

\begin{eqnarray}
k_{GA}(x,y)  &\overset{def}{=} & \sum_{\pi \in A(n,m)} \prod_{i=i}^{|\pi|} e^{-\phi ( x_{\pi_{1}(i)},y_{\pi_{2}(i)} )}  \\ 
& \overset{def}{=}& \sum_{\pi \in A(n,m)} \prod_{i=i}^{|\pi|} \kappa \left( x_{\pi_{1}(i)},y_{\pi_{2}(i)} \right),
\end{eqnarray}
where notation $\kappa=e^{-\phi}$ was introduced for the sake of simplicity. It has been argued that $k_{GA}$ runs over the whole spectrum of the costs and gives rise to a smoother measure than the minimum of the costs, i.e., the DTW distance \cite{cuturi07akernel}. It has been shown in the same paper that $k_{GA}$ is positive definite provided that $\kappa/(1+\kappa)$ is positive definite on $\mathscr{X}$. Furthermore, the computational effort is similar to that of the DTW distance; it is $\mathcal{O}(nm)$. Cuturi argued in \cite{cuturi2011fast} that global alignment kernel induced Gram matrix do not tend to be diagonally dominated as long as the sequences to be compared have similar lengths. 

\begin{figure*}[!ht]
\centering
\subfigure[]{\includegraphics[width=5.5cm]{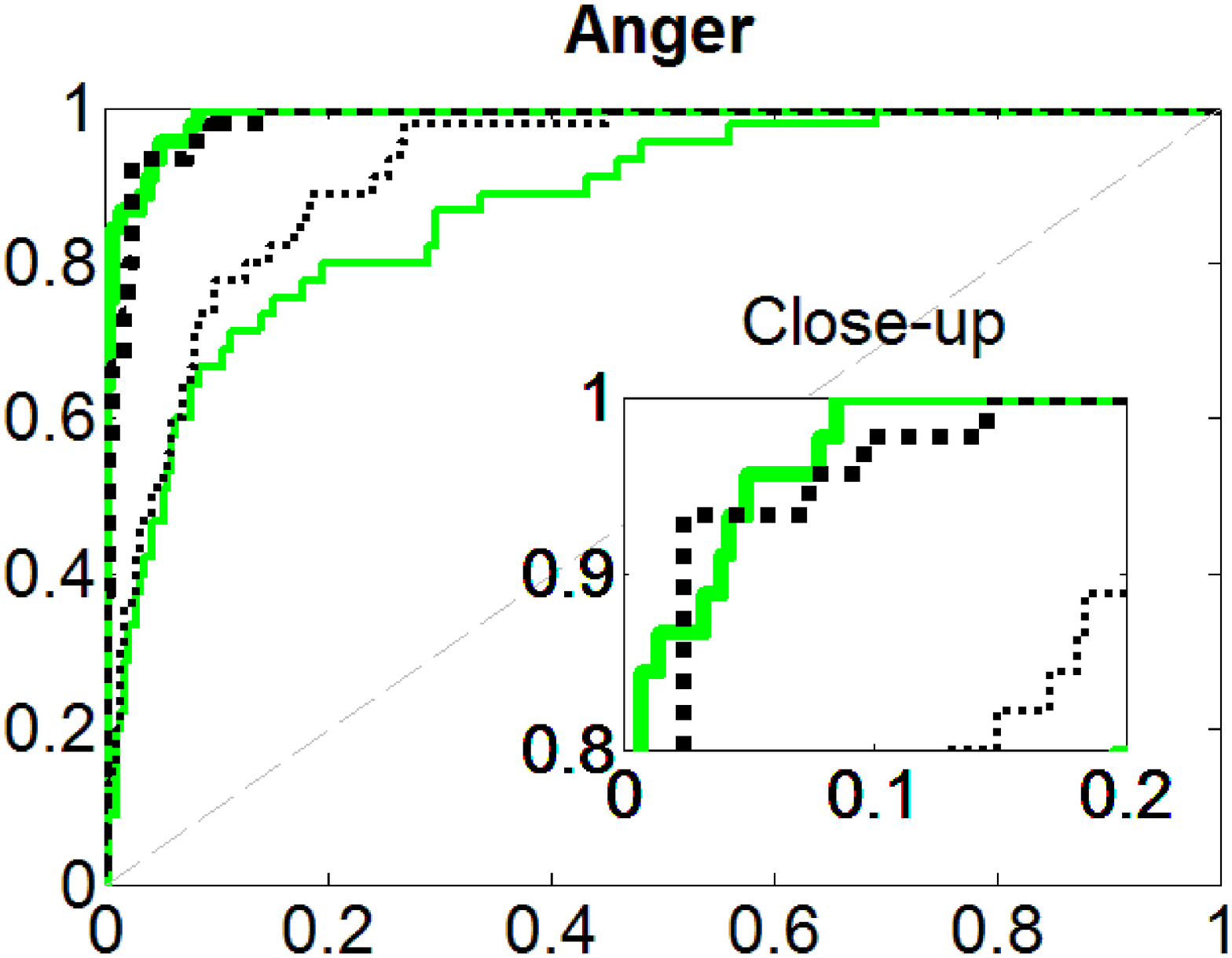}}
\subfigure[]{\includegraphics[width=5.5cm]{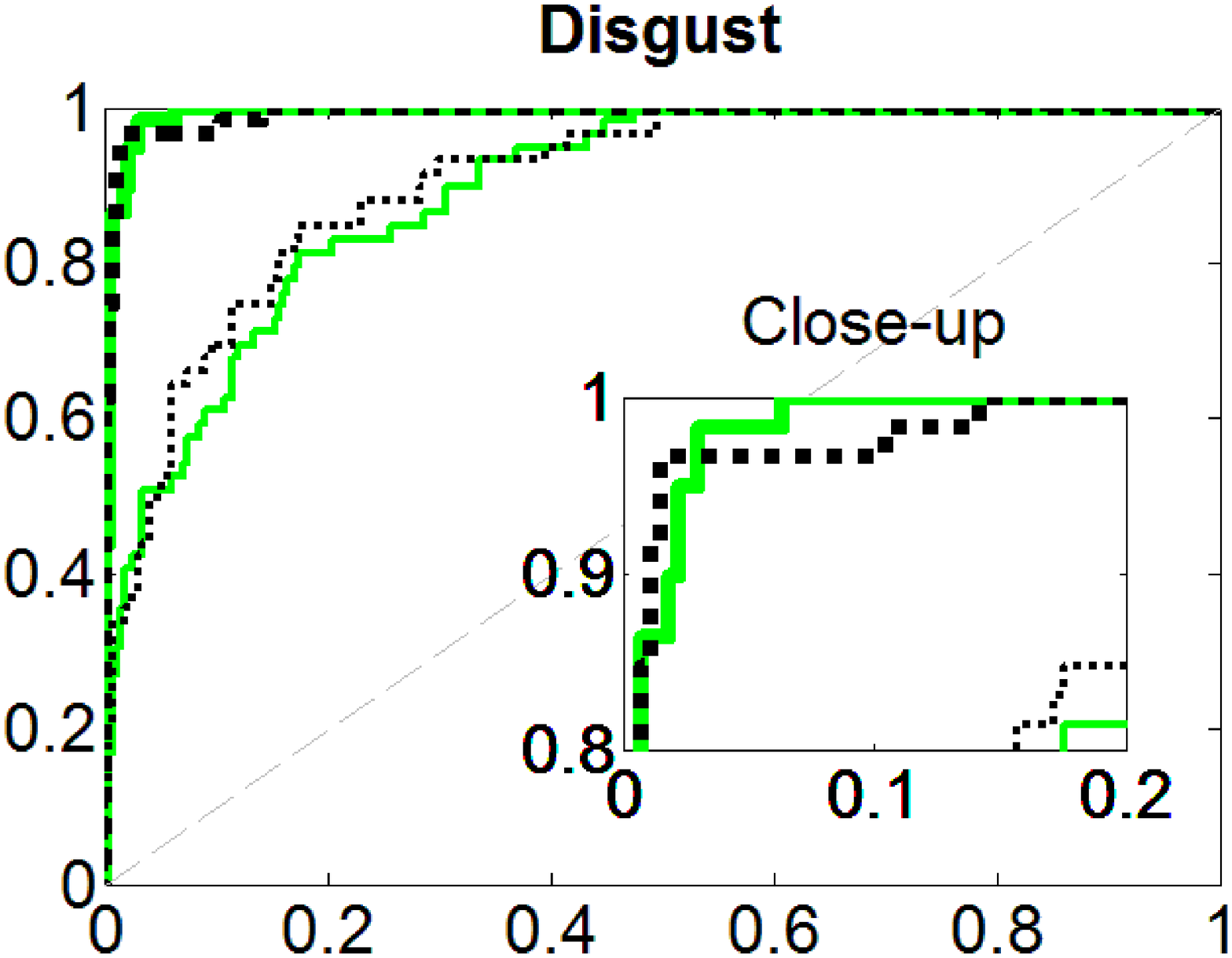}}
\subfigure[]{\includegraphics[width=5.5cm]{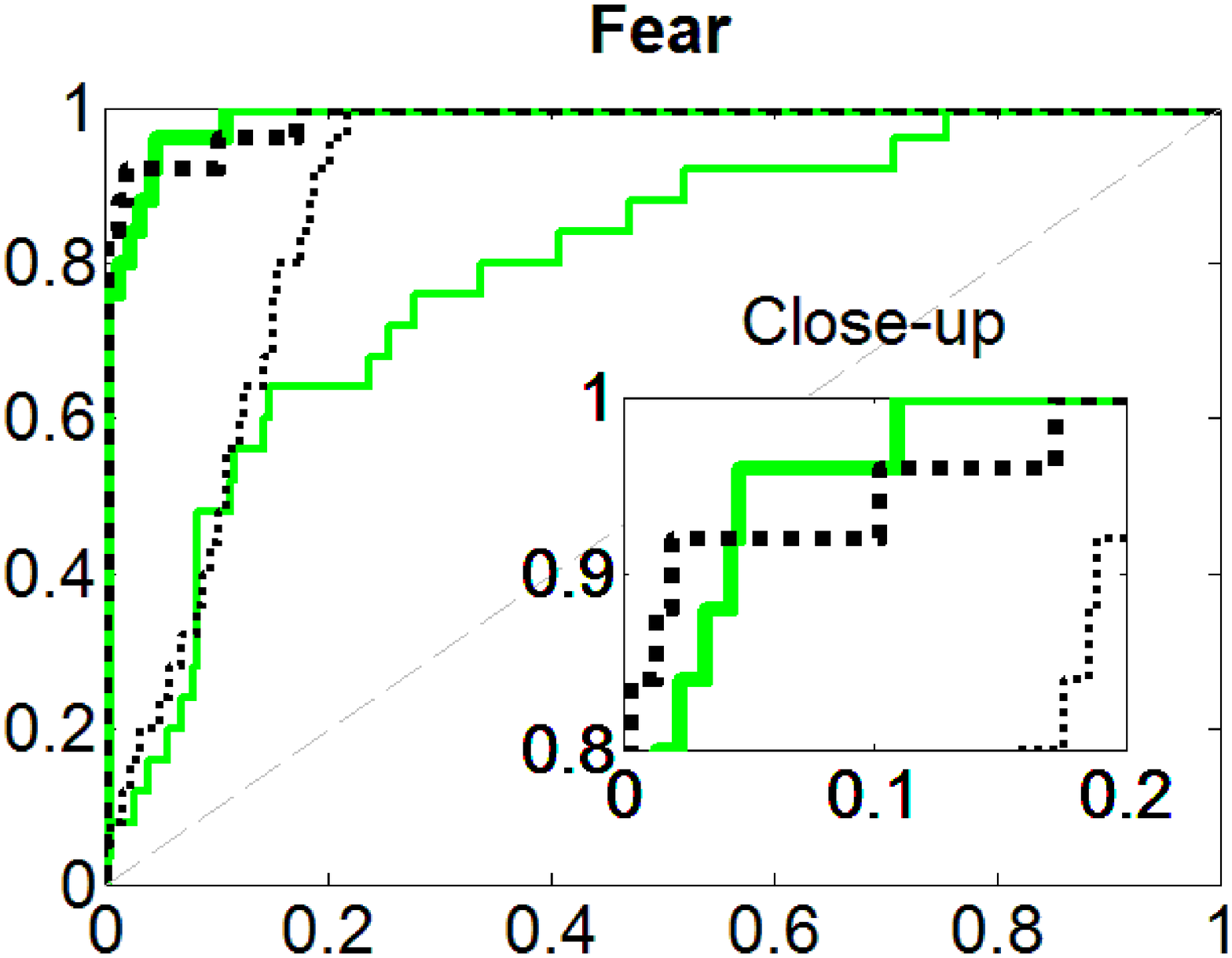}}\hfill
\subfigure[]{\includegraphics[width=5.5cm]{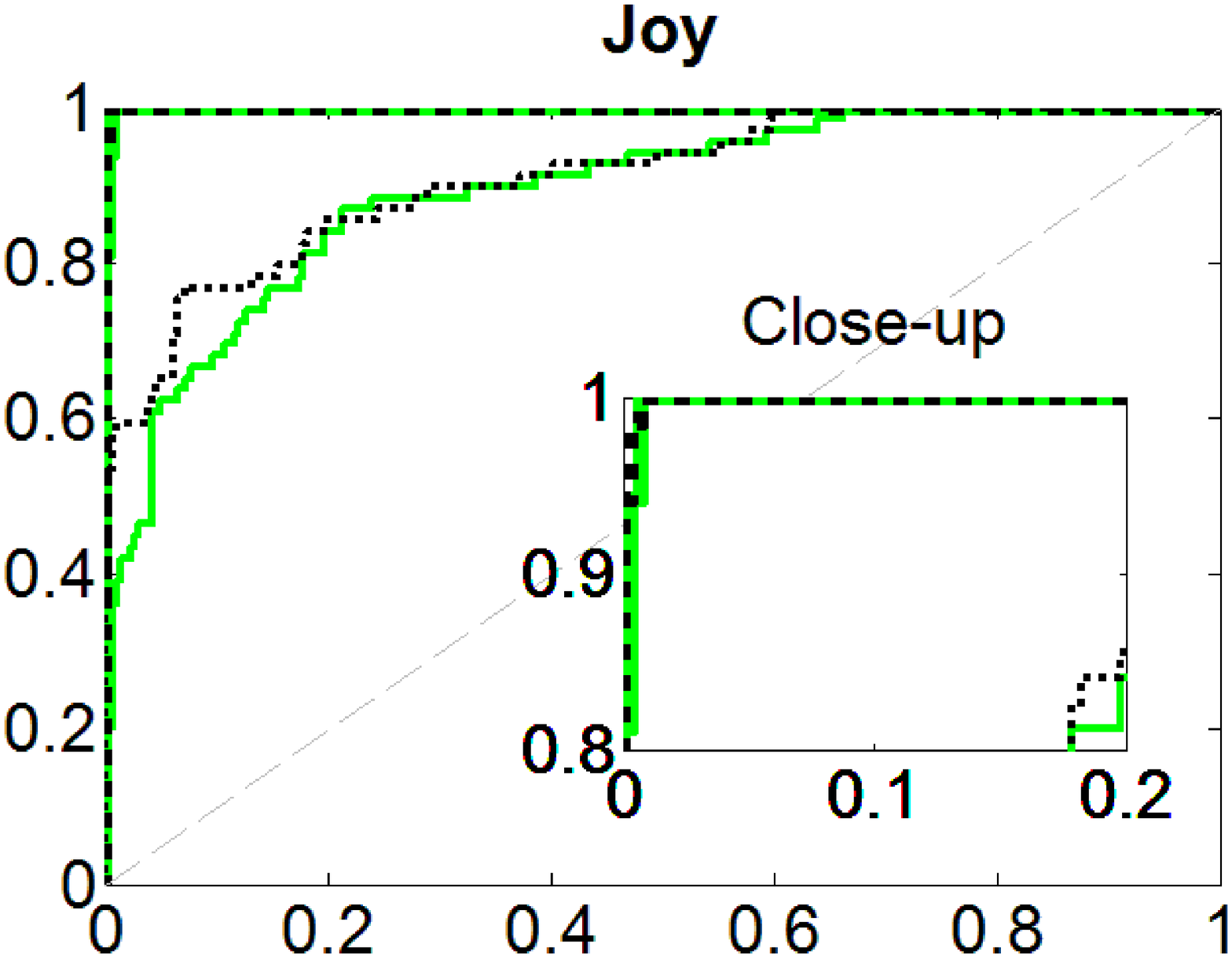}}
\subfigure[]{\includegraphics[width=5.5cm]{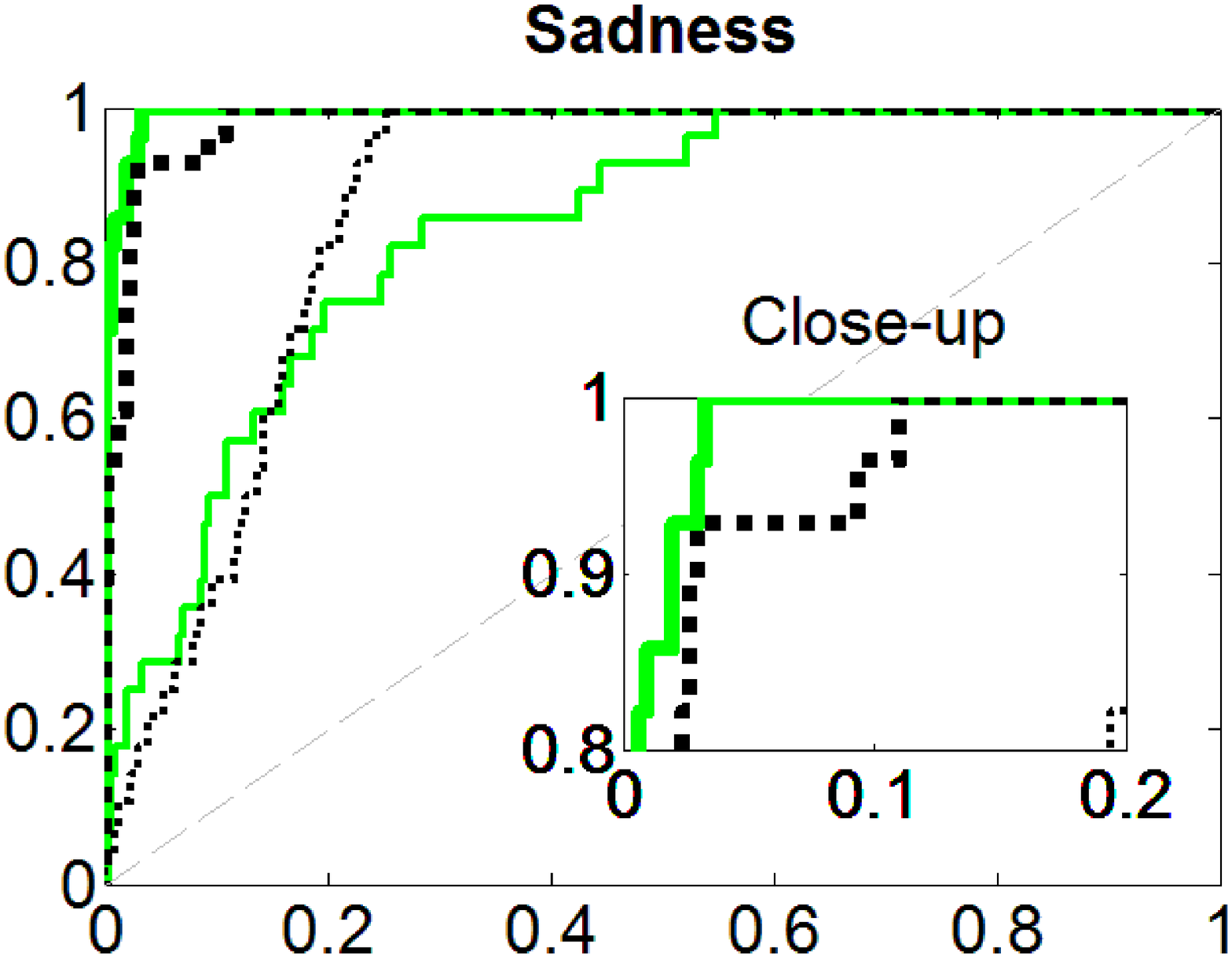}}
\subfigure[]{\includegraphics[width=5.5cm]{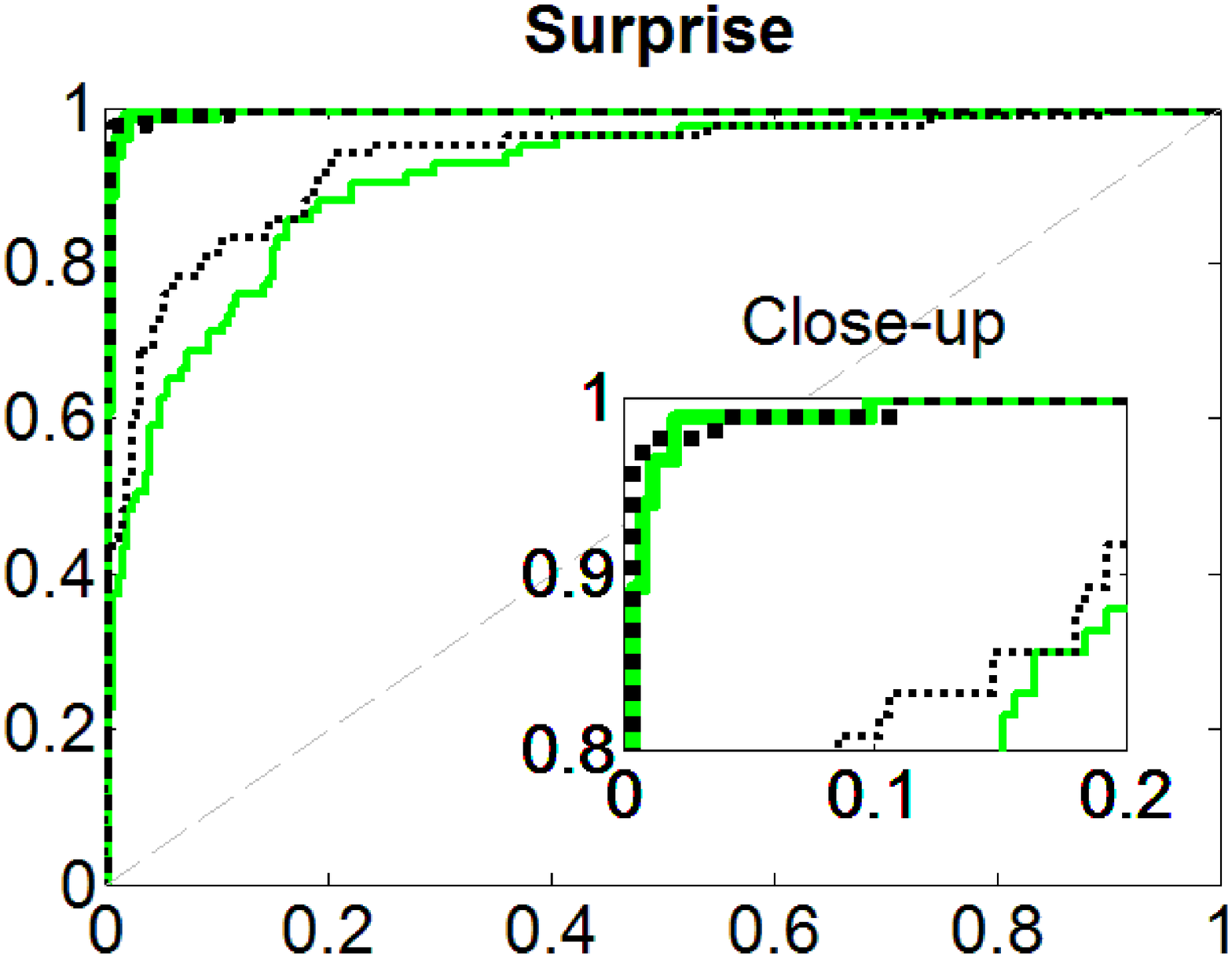}}\hfill
\caption[]{ROC curves of the different emotion classifiers: (a) anger, (b) disgust, (c) fear, (d) joy, (e) sadness and (f) surprise. Thick lines: performance using all frames of the sequences. Thin lines: performance using the 6th frames of the sequences. Solid (dotted) line: results for pseudo DTW (GA) kernel.}
\label{fig:CK_ROC}
\end{figure*}

In our numerical simulations, we used local kernel $e^{-\phi_{\sigma}}$ suggested by Cuturi, where
\begin{equation}
\phi_{\sigma} \overset{def}{=} \frac{1}{2\sigma ^2} ||x-y||^2 + \log\left(2-e^{-\frac{||x-y||^2}{2\sigma ^2}} \right). 
\end{equation}

Fig.~\ref{fig:Gram} (d)-(f) show Gram matrices induced by the GA kernel with different $\sigma$ parameters.

\subsection{Time-series Classification using SVM}

Support Vector Machines (SVMs) are very powerful for binary and multi-class classification as well as for regression problems \cite{CC01a}. They are robust against outliers. For two-class separation, SVM estimates the optimal separating hyper-plane between the two classes by maximizing the margin between the hyper-plane and closest points of the classes. The closest points of the classes are called support vectors; the optimal separating hyper-plane lies at half distance between them.

We are given sample and label pairs $(\b{x}^{(i)},{y}^{(i)})$ with $\b{x}^{(i)}\in\mathbb{R}^m$,  ${y}^{(i)}\in\{-1,1\}$, and $i=1,...,K$. Here, for class '1' and for class '2' ${y}^{(i)}=1$  and ${y}^{(i)}=-1$, respectively. We also have a feature map $\bm{\phi}: \mathbb{R}^m\rightarrow\mathscr{H}$, where $\mathscr{H}$ is a Hilbert-space. 
The kernel implicitly performs the dot product calculations between mapped points: $k(x,y)=\langle\bm{\phi}(x),\bm{\phi}(y)\rangle_{\mathscr{H}}$.
The support vector classification seeks to minimize the cost function

\begin{equation}\label{eq:SVM1}
\min_{\b{w},b,\xi} \frac{1}{2}{\b{w}^T{\b{w}}+C\sum_{i=1}^{K}\xi_i}
\end{equation}

\begin{equation}\label{eq:SVM2}
{y}^{(i)}(\b{w}^T\bm{\phi}(\b{x}^{(i)})+b) \geq 1-\xi_i, \,\,\, \xi_i \geq 0,
\end{equation}
where $\xi_i$ $(i=1, \ldots ,K)$ are the so-called slack variables that generalize the original SVM concept with separating hyper-planes to soft-margin classifiers that have outliers that can not be separated.

\section{Experiments}\label{s:exps}

\subsection{Cohn-Kanade Extended Dataset}

In our simulations we used the Cohn-Kanade Extended Facial Expression (CK+) Database \cite{lucey2010extended}. This database was developed for automated facial image analysis and synthesis and for perceptual studies. The database is widely used to compare the performance of different models. The database contains 123 different subjects and 593 frontal image sequences. From these, 118 subjects are annotated with the seven universal emotions (anger, contempt, disgust, fear, happy, sad and surprise). Action units are also provided with this database for the apex frame.
The original Cohn-Kanade Facial Expression Database distribution \cite{kanade2000comprehensive} had 486 FACS-coded sequences from 97 subjects. CK+ has 593 posed sequences with full FACS coding of the peak frames. A subset of action units were coded for presence or absence.
For these sequences the 3D landmarks and shape parameters were provided by the CLM tracker itself.

\begin{table*}[!ht]
\caption[]{(a) Comparisons with hand-designed spatio-temporal Gabor filters
(Wu et al. 2010 \cite{Tingfan2010}), learned spatio-temporal ICA filters (Long et al.
2012 \cite{Long2012126}) and Sparse Non-negative Matrix Factorization (NMF) filters (Jeni et al. 2013 \cite{jenicontinuous}) on the first 6 frames. (b) Comparisons with boosted dynamic features (Yang et al. 2009 \cite{yang2009boosting}) on the last frames of the sequences.}    
\label{table:CK_comparison}
	\footnotesize 
	\centering
	\subtable[]{
	\tabcolsep=0.11cm
    \begin{tabular}{@{}llllllll@{}}
    \toprule
    Method                            & Anger   & Disg.   & Fear    & Joy     & Sadn.   & Surp.   & Average \\
    \midrule
    Wu et al. \cite{Tingfan2010}      & 0.829   & 0.677   & 0.667   & 0.877   & 0.784   & 0.879   & 0.786   \\
    Long et al. \cite{Long2012126}    & 0.774   & 0.711   & 0.692   & 0.894   & 0.848   & 0.891   & 0.802   \\
    Jeni et al. \cite{jenicontinuous} & 0.817   & 0.908   & 0.774   & 0.938   & 0.865   & 0.886   & 0.865   \\
    This work, DTW                    & 0.873   & 0.893   & 0.793   & 0.892   & 0.843   & 0.909   & 0.867   \\
    This work, GA                     & 0.921   & 0.905   & 0.887   & 0.910   & 0.871   & 0.930   & 0.904   \\
    \bottomrule
    \end{tabular}
    }
    \hfill
    \subtable[]{
    \tabcolsep=0.11cm
    \begin{tabular}{@{}llllllll@{}}
    \toprule    
    Method                              & Anger   & Disg.   & Fear    & Joy     & Sadn.   & Surp.   & Average \\
    \midrule
    Yang et al. \cite{yang2009boosting} & 0.973   & 0.941   & 0.916   & 0.991   & 0.978   & 0.998   & 0.966   \\
    Long et al. \cite{Long2012126}      & 0.933   & 0.988   & 0.964   & 0.993   & 0.991   & 0.999   & 0.978   \\
    Jeni et al. \cite{jenicontinuous}   & 0.989   & 0.998   & 0.977   & 0.998   & 0.994   & 0.994   & 0.992   \\
    This work, DTW                      & 0.991   & 0.994   & 0.987   & 0.999   & 0.995   & 0.996   & 0.994   \\
    This work, GA                       & 0.986   & 0.993   & 0.986   & 1.000   & 0.984   & 0.997   & 0.991   \\
    \bottomrule    
    \end{tabular}    
    }
\end{table*}

\begin{figure*}[!ht]
\centering
\subfigure[]{\includegraphics[width=5.5cm]{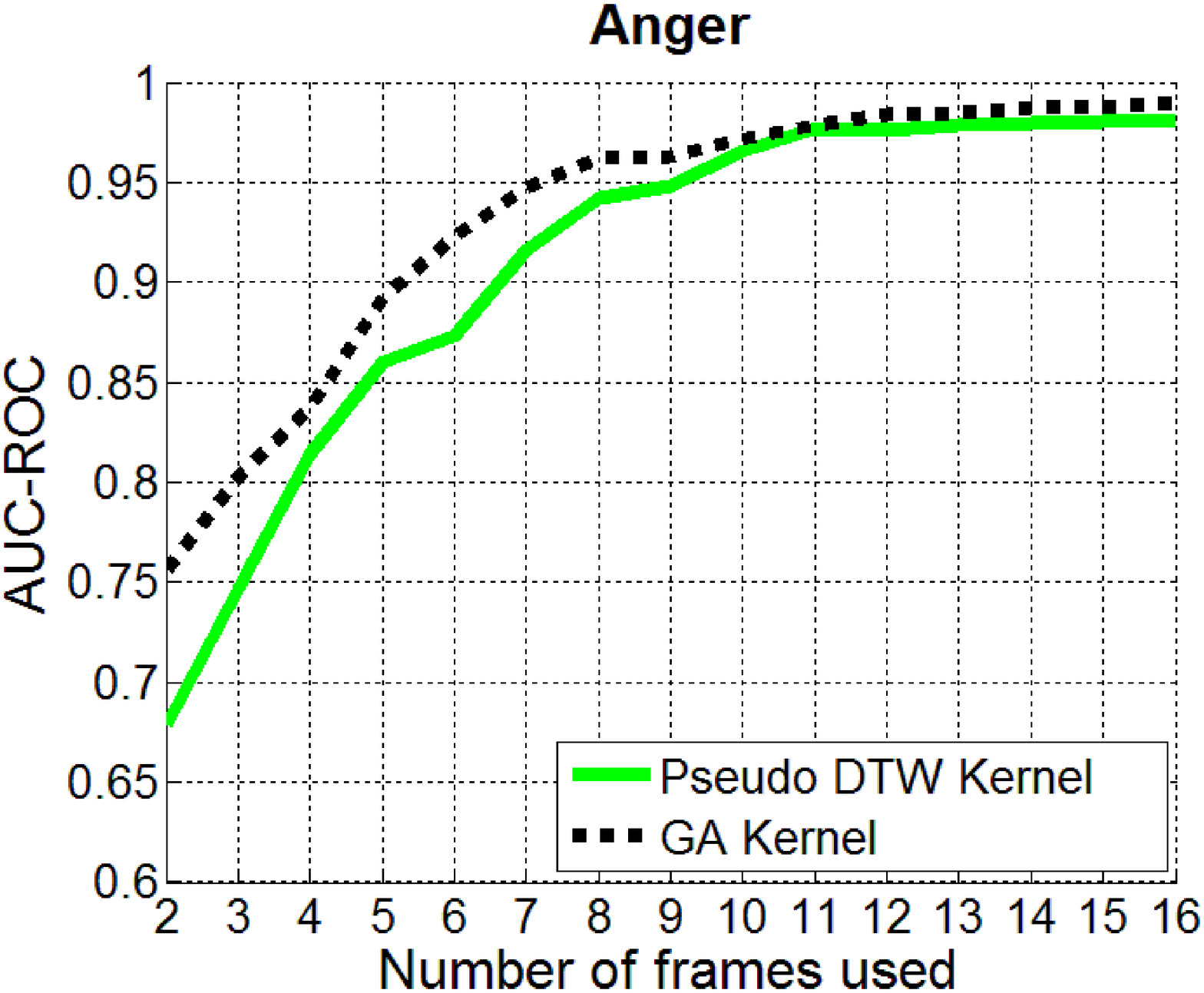}}
\subfigure[]{\includegraphics[width=5.5cm]{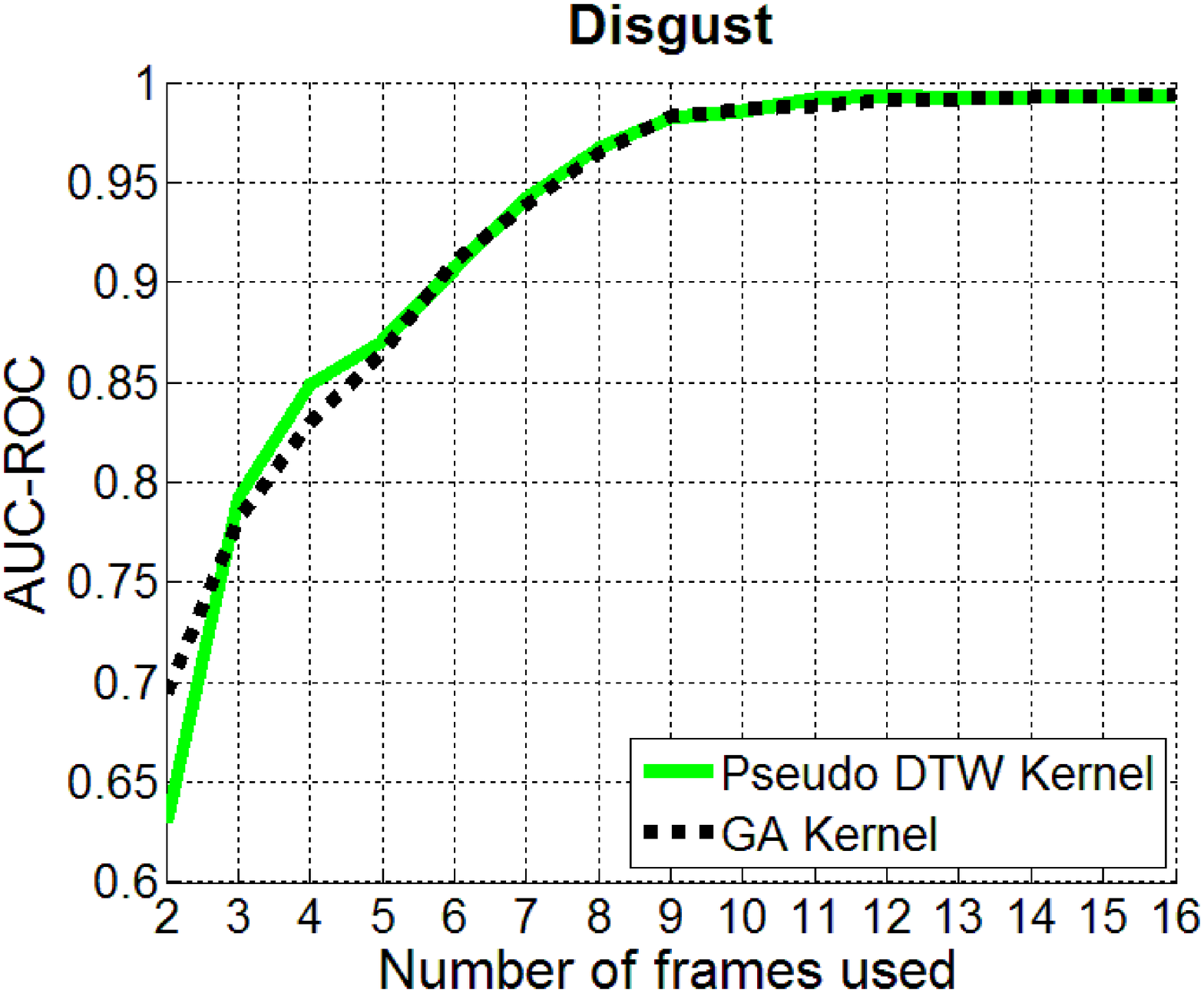}}
\subfigure[]{\includegraphics[width=5.5cm]{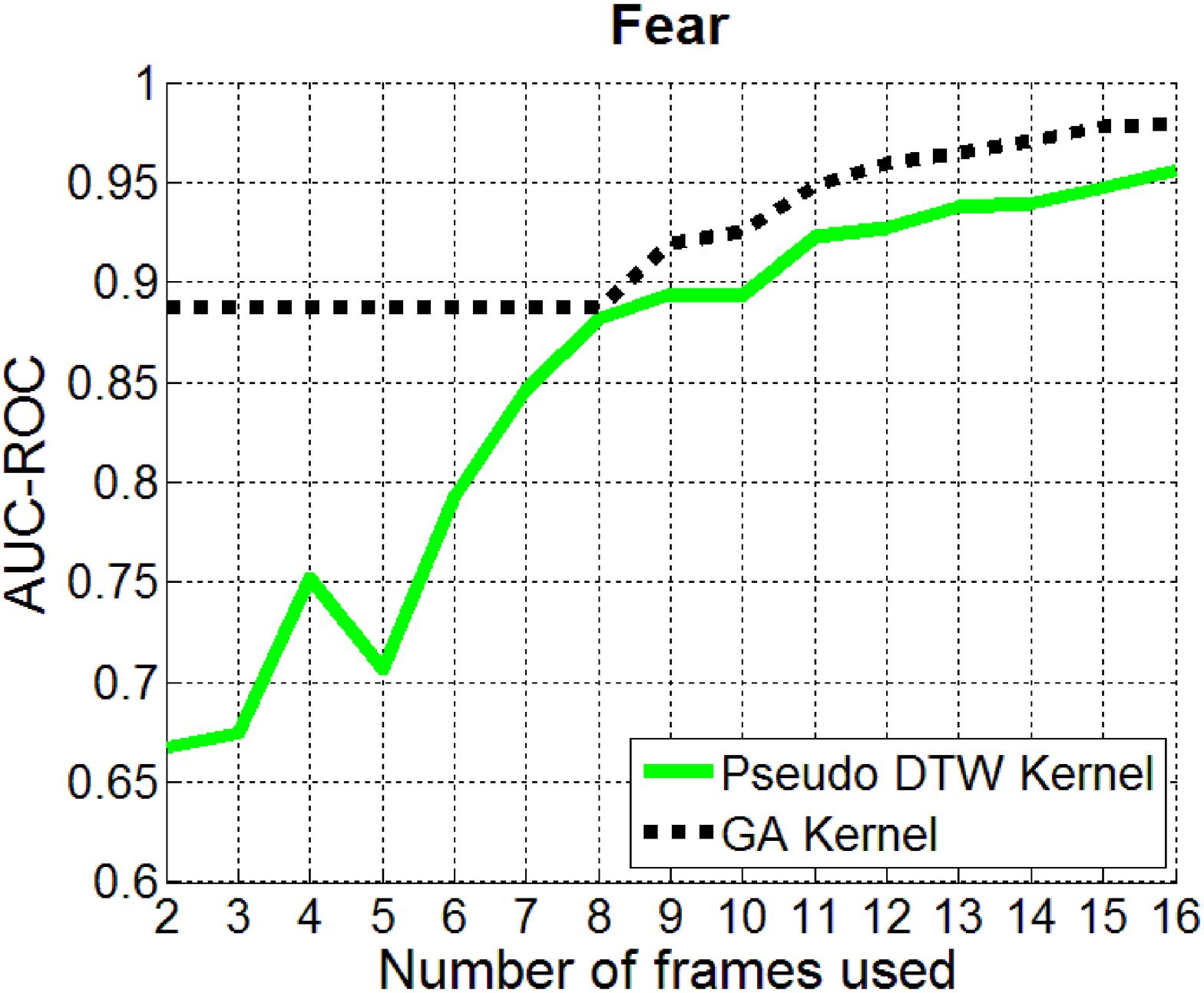}}\hfill
\subfigure[]{\includegraphics[width=5.5cm]{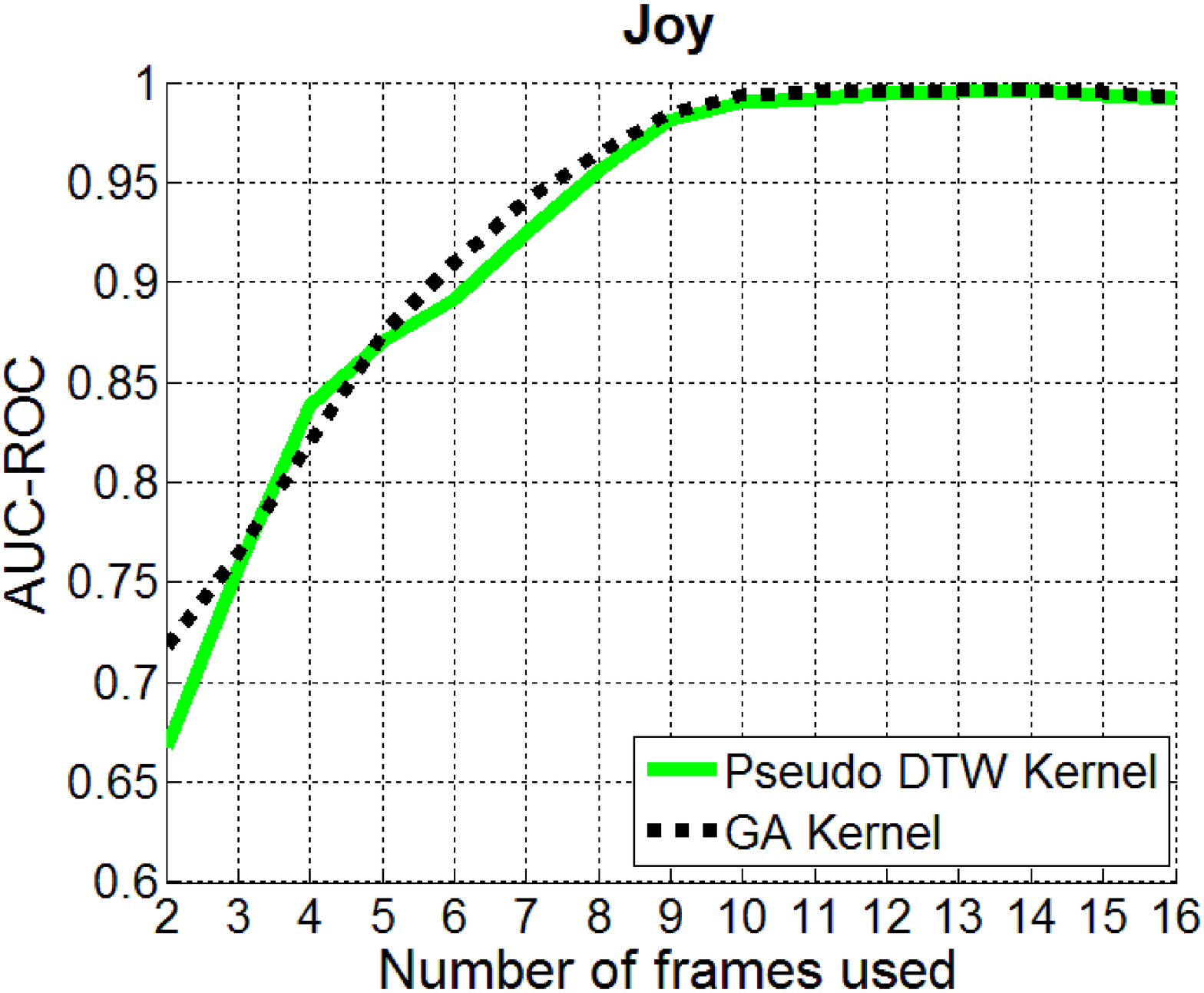}}
\subfigure[]{\includegraphics[width=5.5cm]{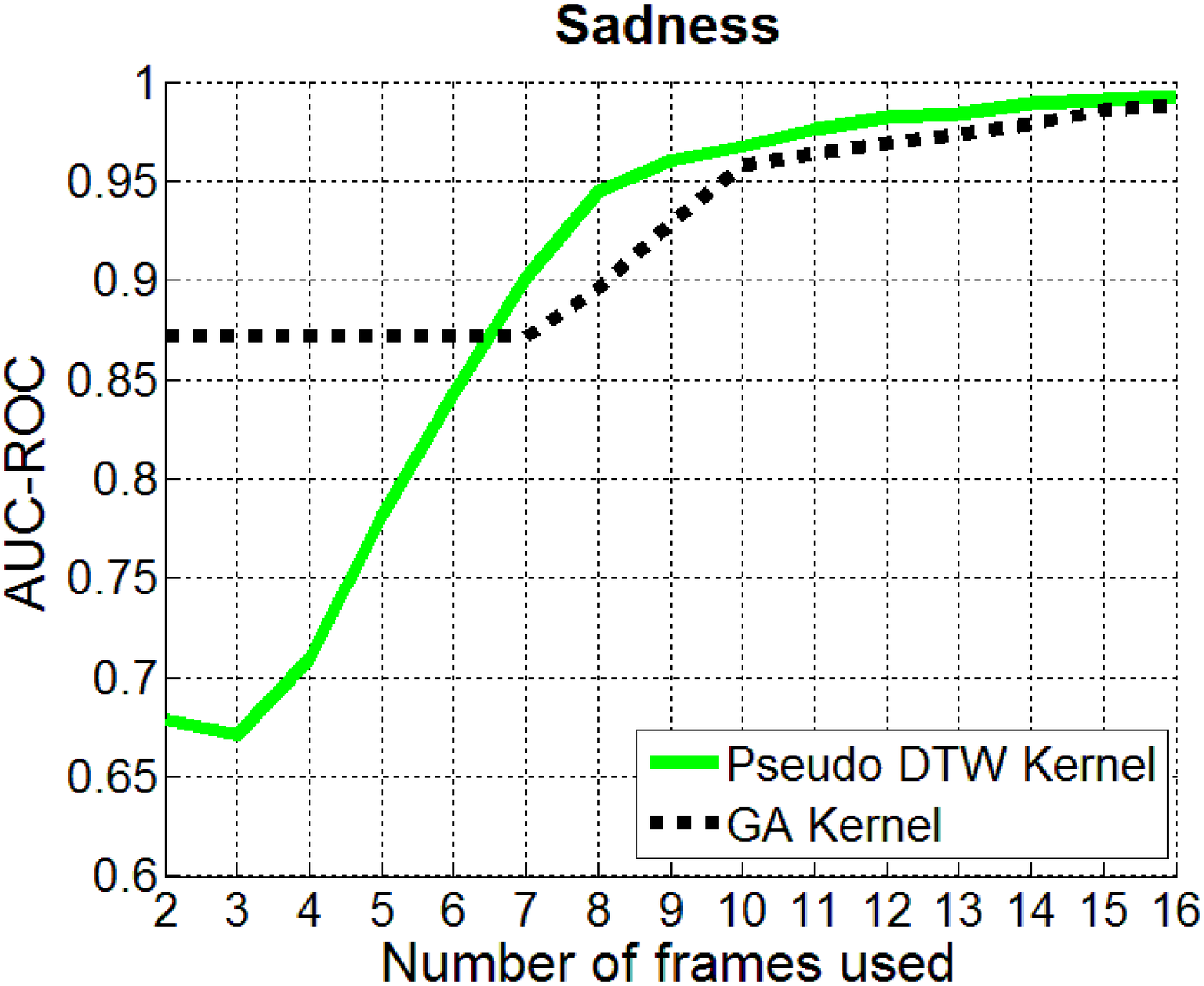}}
\subfigure[]{\includegraphics[width=5.5cm]{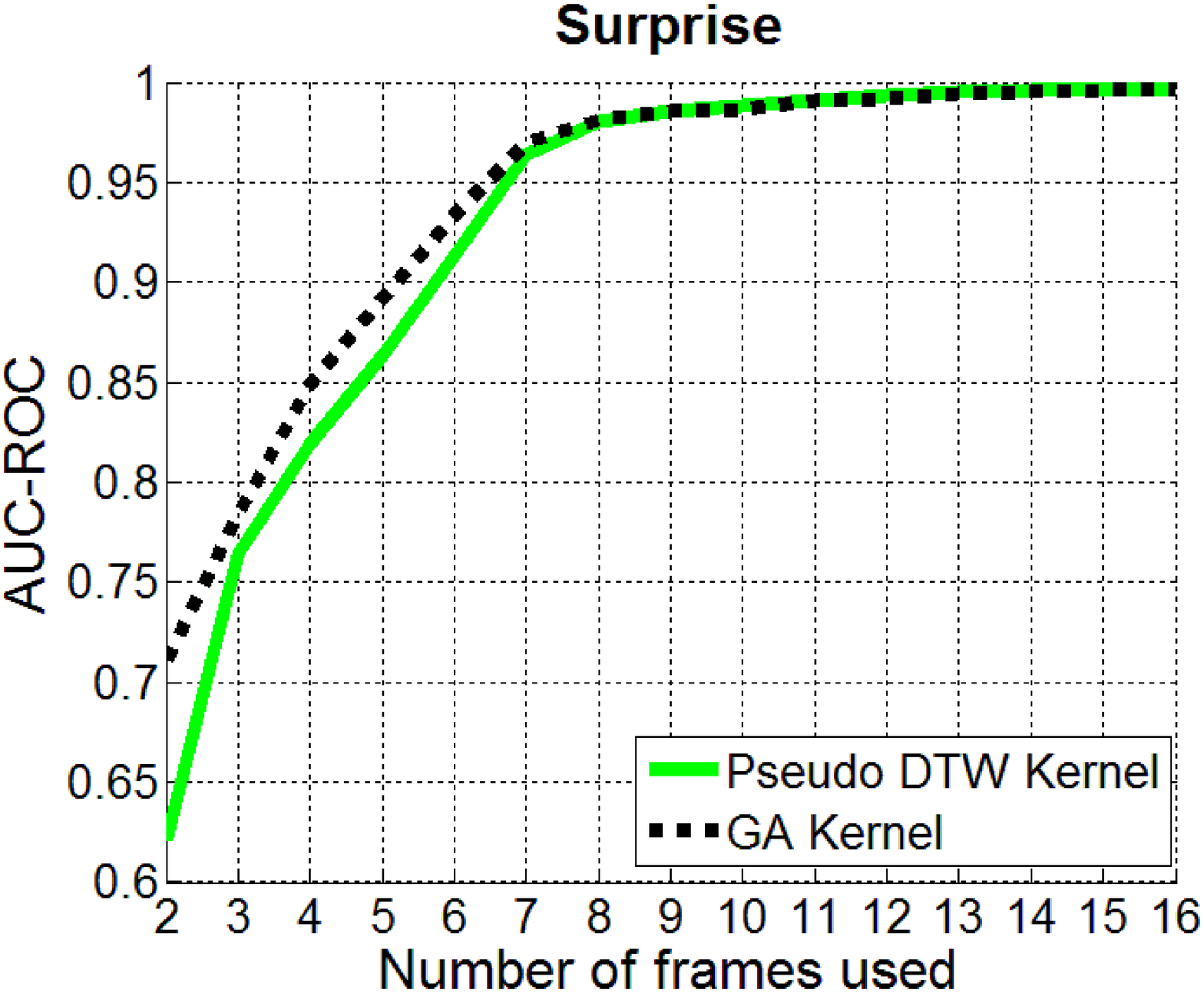}}\hfill
\caption[]{Area Under ROC curve values of the different emotion classifiers: (a) anger, (b) disgust, (c) fear, (d) joy, (e) sadness and (f) surprise. Solid (dotted) line: results for the pseudo DTW (GA) kernel.}
\label{fig:CK_AUCROC}
\end{figure*}

\subsection{Emotional Expression Classification}

In this set of experiment we studied the two kernel methods on the CK+ dataset. We measured the performances of the methods for emotion recognition.

First, we tracked facial expressions with the CLM tracker and annotated all image sequences starting from the neutral expression to the peak of the emotion. The CLM estimates the rigid and non-rigid transformations. We removed the rigid ones from the faces and represented the sequences as multi-dimensional time-series built from the non-rigid shape parameters.

We calculated Gram matrices using the pseudo DTW and the GA kernels and performed leave-one-subject out cross validation to maximally utilize the available set of training data. For both kernels, we searched for the best parameter ($t$ in the case of pseudo-DTW kernel and $\sigma$ in the case of GA kernel) between $2^{-5}$ and $2^{10}$ on a logarithmic scale with equidistant steps and selected 
the parameter having the lowest mean classification error. The SVM regularization parameter ($C$) was searched within  $2^{-5}$ and $2^{5}$ in a similar fashion. If the pseudo-DTW kernel based Gram matrix was not positive semi-definite then we projected it to the nearest positive semi-definite matrix using the alternating projection method of \cite{higham2002computing}.

The result of the classification is shown in Fig.~\ref{fig:CK_ROC}. Performance is nearly 100\% for expressions with large deformations in the facial features, such as disgust, happiness and surprise.

To the best of our knowledge, classification performance with time-series kernels is better than the best available results to date, including spatio-temporal ICA, boosted dynamic features, and non-negative matrix factorization techniques. For detailed comparisons, see Table~\ref{table:CK_comparison}.

\subsection{Early Expression Classification}

Encouraged by the results of the first experiment, we decided to constrain the maximum length of the sequences used in training and testing in order to estimate performance in the early phase of the emotion events. 

We cropped the the time series between 2 and 16 frames and trained kernel SMVs for one-vs-all emotion classification.  Figure~\ref{fig:CK_AUCROC}  shows the classification performance as a function of the maximum length of the sequences. According to the figures, 3-to-4 frames can reach 80\% AUC-ROC performance, whereas 5-to-6 frames are sufficient for about 90\% performance.

\section{Discussion and Summary}

We have studied time-series kernel methods for the analysis of emotional expressions. We used the well known 3D CLM method with the available open source C++ implementation of Jason Saragih. Compared to previous results, we found superior performances both on the first 6 frames and on the last few frames of the sequences collected from the CK+ database. It is notable that we used only shape information and neglected the textural one, since the 3D CLM model can compensate for the head poses making the method robust against head pose variations \cite{Jeni2012785}.

The NMF method \cite{jenicontinuous} that deeply exploits textural features comes close to our method and one expects that mixing the two methods may unite the advantages of the two approaches, namely the robustness of the shape based method against pose variations and light conditions, and sometimes strong textural changes under small landmark position variations. Also, textural changes are less sensitive to the estimation noise of the landmark positions. 

We achieved highly promising results at early times of the series: 3-to-4 frames reached 80\% AUC-ROC performance, whereas 5-to-6 frames were sufficient for about 90\% performance. Such early detection enables timely response in human-computer interactions and collaborations. Furthermore, the early frames of the series have smaller AUC values and should make emotion estimation more robust.

In sum, time-series kernels are very promising for emotion recognition. There is a number of potential improvements to our method, such as (i) joined texture and shape based facial expression recognition using for example, probabilistic SVMs, and (ii) novel DTW optimization methods, like lower bounding or the UCR suite approach \cite{rakthanmanon2012searching} that can make the proposed system tractable for real-time analysis.

\section{Acknowledgments}
The research was carried out as part of the EITKIC\_12-1-2012-0001 project,
which is supported by the Hungarian Government, managed by the National
Development Agency, financed by the Research and Technology Innovation Fund
and was performed in cooperation with the EIT ICT Labs Budapest Associate
Partner Group. (\url{www.ictlabs.elte.hu}) We are grateful to Jason Saragih for providing his CLM code for our work.

{\small
\bibliographystyle{ieee}
\bibliography{AMFG_2013}
}

\end{document}